\documentclass[10pt,twocolumn,letterpaper]{article}

\usepackage{cvpr}
\usepackage{times}
\usepackage{epsfig}
\usepackage{graphicx}
\usepackage{amsmath}
\usepackage{amssymb}
\usepackage{subfig}
\usepackage{url}

\usepackage[breaklinks=true,bookmarks=false]{hyperref}

\cvprfinalcopy % *** Uncomment this line for the final submission

\def\cvprPaperID{****} % *** Enter the CVPR Paper ID here

\begin{document}

%%%%%%%%% TITLE
\title{Semi-Supervised Multitask Learning on Multispectral Satellite Images Using\\ Wasserstein Generative Adversarial Networks (GANs) for Predicting Poverty}

\author{Anthony Perez \thanks{Equal Contributions. Enrolled in CS 231N Convolutional Neural Networks for Visual Recognition during Spring 2017 at Stanford}\\
Stanford University\\
Stanford, CA - 94305\\
{\tt\small aperez8@cs.stanford.edu}
% For a paper whose authors are all at the same institution,
% omit the following lines up until the closing ``}''.
% Additional authors and addresses can be added with ``\and'',
% just like the second author.
% To save space, use either the email address or home page, not both
\and
Swetava Ganguli\footnotemark[1]\\
Stanford University\\
Stanford, CA - 94305\\
{\tt\small swetava@cs.stanford.edu}
\and
Stefano Ermon\thanks{Assistant Professor, Department of Computer Science}\\
Stanford University\\
Stanford, CA - 94305\\
{\tt\small ermon@cs.stanford.edu}
\and
George Azzari\thanks{Research Associate, Department of Earth System Science}\\
Stanford University\\
Stanford, CA - 94305\\
{\tt\small gazzari@stanford.edu}
\and
Marshall Burke\thanks{Assistant Professor, Department of Earth System Science}\\
Stanford University\\
Stanford, CA - 94305\\
{\tt\small mburke@stanford.edu}
\and
David Lobell\thanks{Professor, Department of Earth System Science}\\
Stanford University\\
Stanford, CA - 94305\\
{\tt\small dlobell@stanford.edu}
}

\maketitle
\begin{abstract}
Obtaining reliable data describing local poverty metrics at a granularity that is informative to policy-makers requires expensive and logistically difficult surveys, particularly in the developing world. Not surprisingly, the poverty stricken regions are also the ones which have a high probability of being a war zone, have poor infrastructure and sometimes have governments that do not cooperate with internationally funded development efforts. We train a CNN on free and publicly available daytime satellite images of the African continent from Landsat 7 to build a model for predicting local economic livelihoods. Only 5\% of the satellite images can be associated with labels (which are obtained from DHS Surveys) and thus a semi-supervised approach using a GAN \cite{Salimans16}, albeit with a more stable-to-train flavor of GANs called the Wasserstein GAN regularized with gradient penalty \cite{Gulrajani17} is used. The method of multitask learning is employed to regularize the network and also create an end-to-end model for the prediction of multiple poverty metrics. 
\end{abstract}

%%%%%%%%% BODY TEXT
\section{Introduction}
In a recent paper \cite{Jean16}, Convolutional Neural Networks (CNNs) were trained to extract image features from daytime satellite images with nightlight luminosity labels from all over the African continent which were then shown to explain upto 75\% of the variation in local economic livelihood indicators. This is a significant step towards using free and publicly available satellite imagery to measure the need for or the impacts of economic and social policies without requiring expensive and logistically difficult surveys. This ability would specifically benefit policy makers or philanthropic organizations to direct aid and effort in places that most need them \cite{un15}, \cite{devarajan13}, \cite{jerven13}. If we focus on just the continent of Africa, gaps in collection of data are exacerbated by poor infrastructure, ravaging civil wars, and inadequate government resourcing. According to data from the World Bank, 39 of 59 African countries conducted fewer than two surveys during the years 2000 to 2010, from which nationally representative poverty measures could be constructed. Of these, 14 conducted no such surveys during this period \cite{un15}. Sometimes, governments do not have enough resources to conduct these surveys while others lack the motivation to highlight their lackluster performance; least of all document it \cite{devarajan13}, \cite{sg15}. Clearly, closing these large data gaps for accurate measurements of the targets of the United Nations Sustainable Development Goals in every country is prohibitively expensive and institutionally difficult \cite{jerven13}. 

Recently, an increasing and consistent supply of satellite data has added accurate and sophisticated geo-spatial information into studying environmental health, economics, geo-politics, etc. \cite{planet2015pres}. The unique insight attainable from such datasets has continued to be integrated into the intelligence operations of both public (e.g. NSA, CIA) and private sector (e.g. Skybox, etc.) entities to the point that a number of modern processes would not be possible in their absence \cite{meyer2014spysats}. With advances in sensor technology, modern satellites are capable of capturing hyperspectral data which enables the use of satellite data in a variety of ways which were not thought to have been possible before \cite{romero2014unsupervised}. Inspired from \cite{Jean16}, we train CNNs to identify satellite image features that are used to predict an \textit{asset wealth index} (AWI, as defined in \cite{Jean16}) that is a stable measure of a household's long-run economic status and serves as a proxy quantitative measure for the United Nations Sustainable Development Goals in continental Africa. As part of the Sustainability and AI Lab at Stanford, we want to further this effort by taking advantage of recent progress in semi-supervised learning using Wasserstein GANs. We collected multi-spectral imagery from Landsat 7 covering the African continent ranging from 2004 to 2015 (amounting to about 22 TB). As labels, we have access to Demographic and Health Surveys (DHS) which provide the AWIs for multiple countries for the years of interest. However, the coverage of these surveys is sparse compared to the large amounts of satellite image data that we have at our disposal. Therefore, we choose to implement semi-supervised learning algorithms that can utilize large amounts of unlabelled data along with a small set of labeled data. \cite{Salimans16} and \cite{Kuleshov17} demonstrate methods of using multi-conditional likelihood objectives and shared latent variables for variational inference to boost prediction on domains where labels are scarce but data is abundant, which is the setting under which our project operates. \cite{Salimans16} demonstrates that GANs \cite{Goodfellow14} can be used to take advantage of unlabeled data and thus perform semi-supervised learning. However, GANs can be finicky and are in general difficult to train. A variant of GANs, called the Wasserstein GAN (GAN) has been shown to be more stable to train \cite{Arjovsky17}. WGANs can sometimes generate low-quality samples or fail to converge in some settings due to the use of weight clipping to enforce a Lipschitz constraint on the critic. This problem is alleviated by the use of a gradient penalty \cite{Gulrajani17}. We use the WGAN-GP approach in tandem with the approach of \cite{Salimans16} to build our semi-supervised network. In addition, \cite{Caruana97} provides evidence that multi-task learning improves generalization of machine learning models ``by using the domain information contained in the training signals of related tasks as an inductive bias". Instead of building an individual model for each poverty metric of interest to us, we resort to multitask learning with the methods proposed in \cite{Kuleshov17} by setting up each poverty metric prediction task as a classification task at the end of the discriminator network and use the available Landsat images to predict AWI across countries in Africa. This enables us to learn the correlations between these metrics as well as regularize the large capacity network. The details of this procedure are described in the sections below. The input to our model is a satellite image and the output is the predicted class in each classification task which in turn is our prediction range of a particular poverty metric.  For certain metrics, policy makers seldom care about the actual value of the metric. Rather, the range is more important in policy making. For example, ``between 10 to 20 people suffer from malnutrition" is as informative as ``12.83 people suffer from malnutrition". This justifies our approach of setting up each prediction task as a classification task. The model, consisting of the generator, discriminator and each successive classification task is trained end-to-end. The prediction tasks that are of interest to us are to predict nightlight intensity, population density, distance to nearest road, land cover type, and AWI simultaneously. Of these, predicting AWI is of utmost importance. In addition to the validation accuracies obtained from the discriminator, models are also compared based on a metric measuring accuracy on predicting this score. Once the model is trained, image features are extracted from the last layer of the discriminator and a cross-validated linear model is built on top of it. A quantitative measure of the performance of our models is the Pearson Correlation Coefficient between the predicted AWI and the ground truth available from the DHS surveys. This is consistent with the approach employed in \cite{Jean16}. Multiple experiments, each described in subsequent sections, are performed to obtain a model that best predicts the AWI. Qualitatively, we also visualize the first layer weights from the discriminator to understand the sensitivity of the initialization schemes employed in the experiments to the prediction accuracy. 

\section{Satellite Images for Related Work}
Use of passively collected data for predicting economic or social indicators or to study natural disasters and their local effects has been a growing trend in the research community. Sources of such data could be satellites, social media, mobile phone networks, etc. A popular recent approach uses satellite images of luminosity at night (``nightlights") to estimate economic activity \cite{hsw12}, \cite{cn11}, \cite{mp13} and \cite{ps16}. \cite{nelson1997roads} use satellite images to predict land use while \cite{dwyer1998global} propose a method for spatial and temporal analysis of life-threatening vegetation fires using satellite images that may be useful for weather departments and government agencies to create contingency plans during wildfire season. In addition, using convolutional neural networks and GANs on geospatial data in unsupervised or semi-supervised settings has also been of interest recently; especially in domains such as food security, cybersecurity, satellite tasking, etc. (\cite{dunnmon2019predicting,ganguli2019predicting,ganguli2019geogan}). \cite{doll2006mapping} and \cite{sutton2002global} use satellite images to study local economic activities and their correlation to global economic indicators like Gross Domestic Product (GDP), Ecosystem Services Product (ESP), etc. The economic and ecological costs of deforestation in Africa and South America and its effects on the local markets has been investigated from a scientific and econometrics viewpoint in \cite{omo2011mangrove} and \cite{munoz2008paying}. As developing countries urbanize and strain the availability of resources, studying the environmental and economic effects of this rapid change is aided by temporal analysis of satellite imagery. \cite{carlson2000impact} and \cite{owen1998assessment} study the impact of urbanization on micro-climates of regions surrounding these cities while \cite{seto2011meta} are able to measure the land use due to urban expansion. Wall Street firms also use satellite surveillance to gather market-moving information. Private-sector companies like DigitalGlobe in Colorado and GeoEye in Virginia build and launch satellites and take pictures for US government intelligence agency clients and private-sector satellite analysis firms like Remote Sensing Metrics LLC. Finally, satellite images have also been used to track utilization of renewable energy sources by \cite{nygaard2010using} and \cite{azhari2008new}.

\section{Description of Data and Preprocessing} \label{section:Data}
As mentioned in the introduction, we collected multi-spectral imagery from Landsat 7 (Enhanced Thematic Mapper Plus (ETM+)) \cite{landsat} covering the African continent ranging from 2004 to 2015 (amounting to about 22 TB). Images from this satellite have 9 spectral bands viz. (i) Blue, (ii) Green, (iii) Red, (iv) Near Infrared (NIR), (v) Shortwave Infrared (SWIR) 1, (vi) Shortwave Infrared (SWIR) 2, (vii) Thermal 1, (viii) Thermal 2 and (ix) Panchromatic. The spatial resolution of bands 1 to 6 is 30m while the resolution of bands 7 and 8 is 60m and that of the last band is 15m. All bands can collect one of two gain settings (high or low) for increased radiometric sensitivity and dynamic range, while Band 6 (Thermal) collects both high and low gain for all scenes and is thus provided as two separate bands. The approximate scene size is 170 km north-south by 183 km east-west (106 mi by 114 mi). Rather than using the Blue, Green, and Red bands as provided by Landsat 7, these visible bands are first pan-sharpened \cite{pansharpening_paper} \cite{pansharpenening_website} which results in 15m resolution versions of the color bands.  Pan-sharpening has proved to be beneficial in a variety of satellite imagery tasks \cite{pansharpening_paper} which justifies its usage.

\begin{figure}[h]
    \centering
    \subfloat[][]{
    	\includegraphics[width=0.31\linewidth]{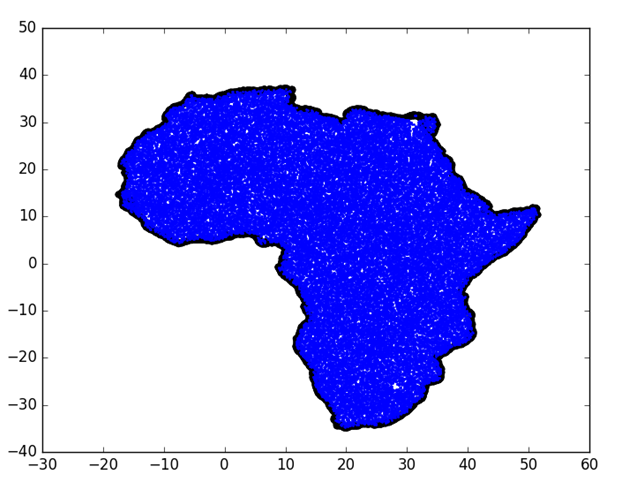}
    } \hfill
    \subfloat[][]{
    	\includegraphics[width=0.31\linewidth]{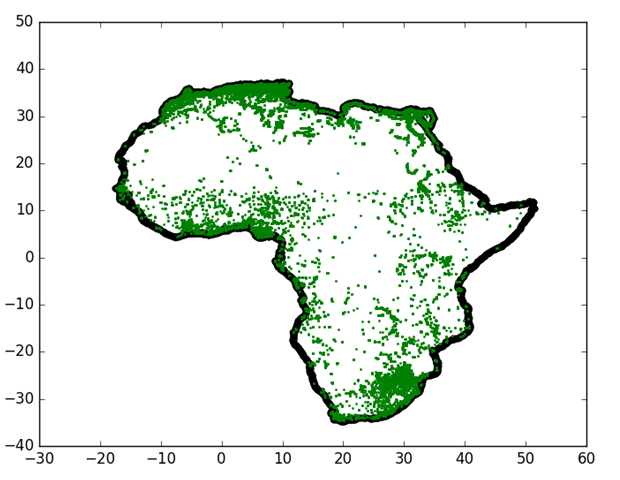}
    } \hfill
    \subfloat[][]{
    	\includegraphics[width=0.31\linewidth]{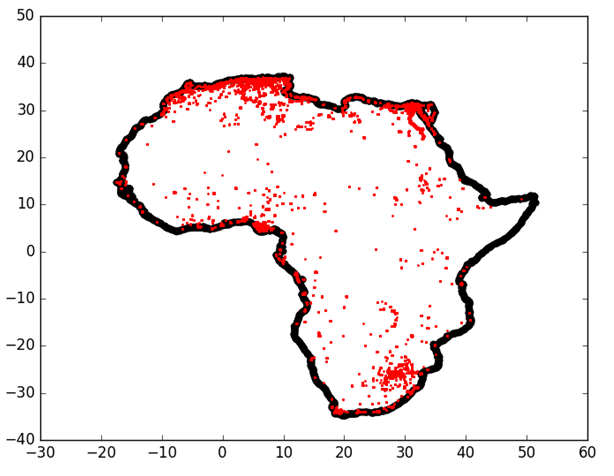}
    } \hfill
    \caption{Geographic locations of satellite image samples which are classified as rural (class 0 in blue), semi-urban (class 1 in green), and urban (class 2 in red) when sampling is performed from the entire African continent.}
    \label{old_locs}
\end{figure}

\begin{figure}[h]
    \centering
    \includegraphics[width=0.3\textwidth]{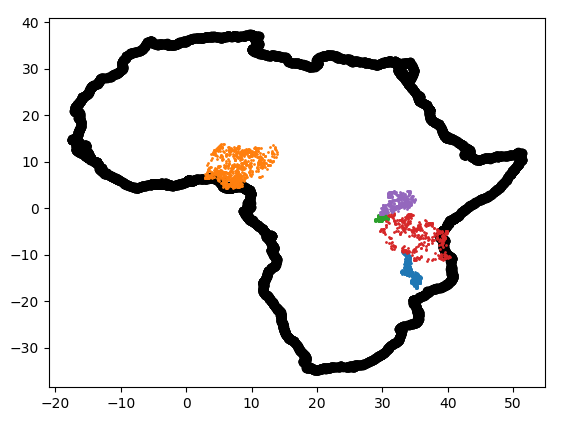}
    \caption{The locations of the most recent DHS survey in each country used in by \cite{Jean16}.  There are 4839 locations in total.}
    \label{fig:dhs_survey_locations}
\end{figure}

As labels, we have access to Demographic and Health Surveys (DHS) which provide the AWIs for multiple countries in Africa for the years of interest. However, the coverage these surveys provided is sparse compared to the large amounts of satellite image data that we have at our disposal. Figure \ref{fig:dhs_survey_locations} shows the locations in Africa where DHS labels are available. While sampling training data, which is a mix of labeled and unlabeled data, we experiment with two sampling strategies. One way to sample is to sample uniformly all over the African continent. Shown in figure~\ref{old_locs} are the geographic locations of the available satellite data from Landsat that are classified as rural, semi-urban, and urban when this sampling strategy is used. The figures are slightly deceptive, as a single pixel corresponds to an area much larger than an image. The data is class balanced, with 50,000 images extracted for each class.  However, this comes at a cost of overlapping in the imagery for the semi-urban and urban classes.  Furthermore, very few locations that have satellite images associated with them have DHS AWI labels.
%\vspace{-1cm}
\begin{figure}[h]
    \centering
    \subfloat[][]{
    	\includegraphics[width=0.31\linewidth]{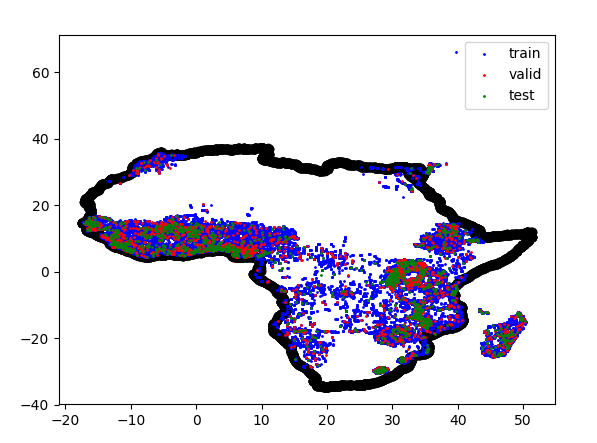}
    } \hfill
    \subfloat[][]{
    	\includegraphics[width=0.31\linewidth]{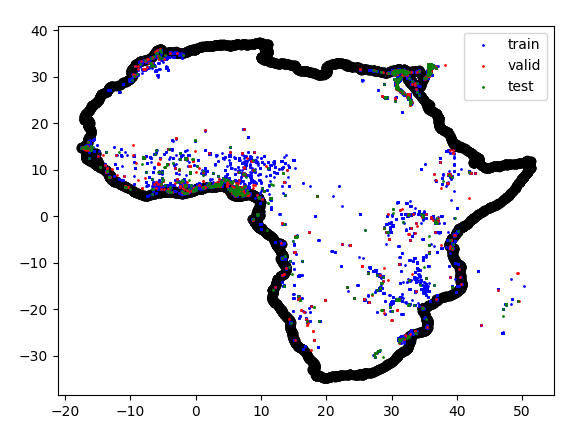}
    } \hfill
    \subfloat[][]{
    	\includegraphics[width=0.31\linewidth]{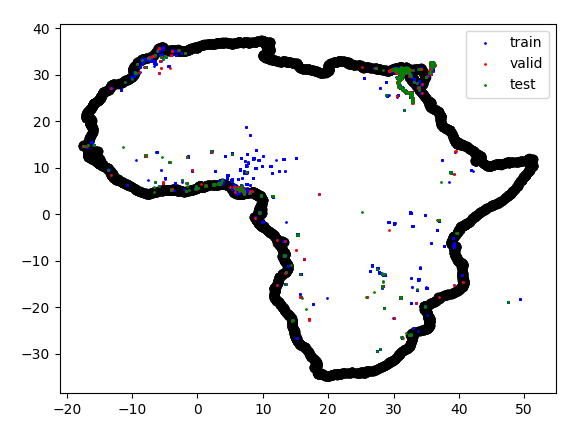}
    } \hfill
    \caption{Geographic locations of satellite image samples which are classified as (a) rural (class 0), (b) semi-urban (class 1), and (c) urban (class 2) when the sampling is performed from regions that are near the locations where DHS survey labels are available. Within each class, the data is split as training (blue), validation (red) and test (green)}
    \label{new_locs}
\end{figure}
%\vspace{-0.8cm}
\begin{figure}[h]
    \centering
    \includegraphics[width=0.8\linewidth]{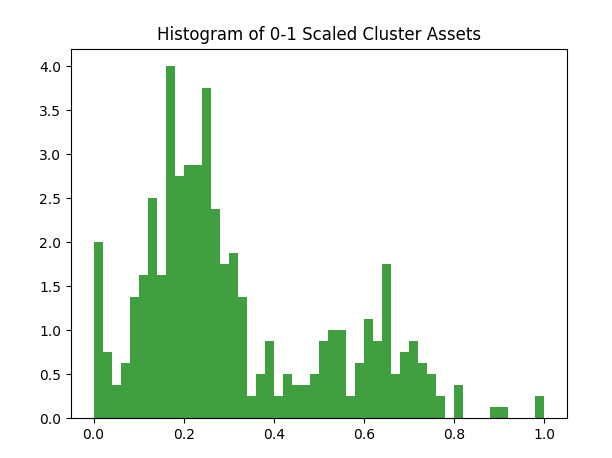}
    \caption{A histogram of scaled assets with 50 buckets demonstrating the distribution of the assets. Clearly, 50 bins is too large since many buckets are left empty. For this metric, an appropriate number of bins is 30.}
    \label{fig:asset_histogram}
\end{figure}

To alleviate this problem, another strategy is to restrict the distance of samples from the locations where labeled data is available. This proximity is an assurance that the labeled and unlabeled images come from nearly similar image distributions. Shown in figure~\ref{new_locs} are the geographic locations of the satellite image samples when sampling is done from regions close to the locations where DHS labels are available following the second sampling strategy. 
\begin{table}
\centering
\begin{tabular}{ |c|c|c|c| } 
 \hline
 Class      & 0     & 1     & 2     \\  \hline
 Train      & 59896 & 18486 & 13140 \\ 
 Validation & 2805  & 883   & 788   \\ 
 Test       & 1330  & 681   & 711   \\  \hline
 Total      & 64031 & 20050 & 14639 \\
 \hline
\end{tabular}
\caption{A description of the per-class distribution of the available data and a 70-20-10 train-val-test split \label{table_data}}
\end{table}
Transitioning to the Africa-wide multitask setting, we have 91522 total satellite images which are bucketed into the 3 classes mentioned before i.e. rural, semi-urban and urban. As described in table~\ref{table_data}, there are 64031 images in the rural bucket, 20050 images in the semi-urban bucket and 14639 images in the urban bucket. As is evident, these classes are not class-balanced. The train-val-test splits are created to ensure that they do not overlap with each other. However, there are images within the training dataset that have partial overlaps. Each task in the multitask framework is set up as a classification task. Since the final predictions for each task of interest to us corresponds to a continuous value, we bin the range of the values of each task into an appropriate number of buckets. If the number of buckets is too low, then we risk having many data points in few of the buckets which results in losing the ability to differentiate between them. If the number of buckets is too high, some buckets may be empty and have no data points representing them. As a case study, the AWI is binned into 50 buckets in figure ~\ref{fig:asset_histogram}. Clearly, this is the case where the number of buckets is too high and some buckets are completely empty. The appropriate number of buckets for AWI turns out to be 30. 

\begin{figure}[h]
    \centering
    \includegraphics[width=0.8\linewidth]{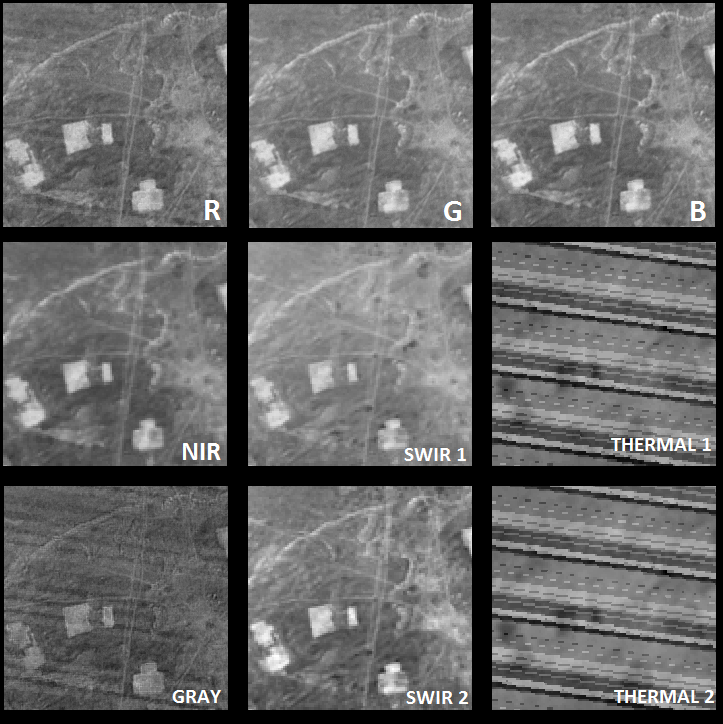}
    \includegraphics[width=0.8\linewidth]{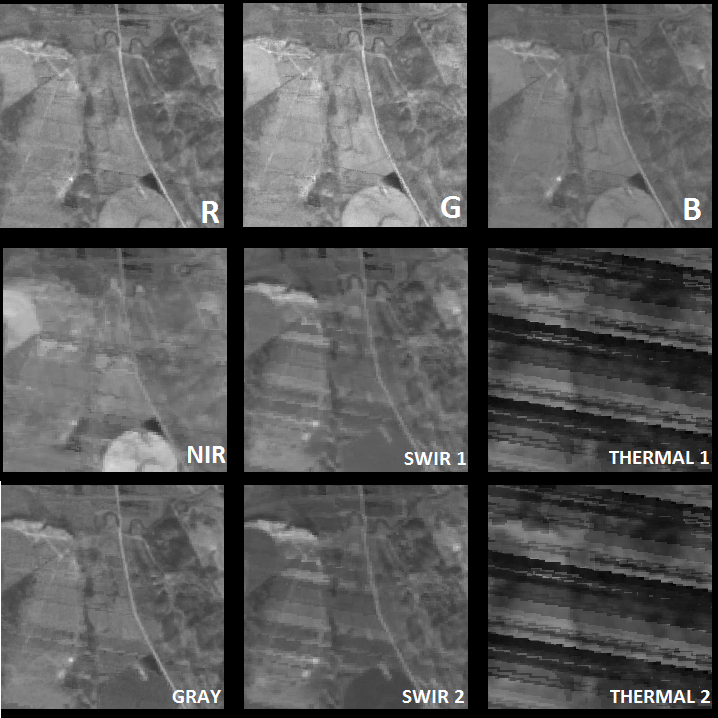}
    \caption{Two examples of actual Landsat 7 images and their 9 spectral bands.\label{fig:real_images}}
\end{figure}

Each modeling task performed in the multi-task training paradigm is associated with an ``importance weight" that is proportional to the importance of accuracy of prediction for that task. Said differently, the weight is proportional to the cost incurred when an incorrect prediction is made for a particular learning task. For example, in our case being able to predict the nightlight intensity and AWI score correctly is more important than some of the other tasks. These importance weights are supplied from collaboration with economic and policy experts in the lab. In addition, every example in a class is weighted inversely proportional to the number of available examples in that class so that the total importance of all examples per class is the same. This ensures that the model does not selectively do well in predicting a certain class while failing to predict another class. Thus, in effect, we calculate the product of the two aforementioned weights as a ``weight-per-example-per-class-per-task''. We have 4839 locations where we have labels from the DHS survey for AWI values and 86683 locations where there are no labels for the asset values. We call this dataset the ``real dataset" and distinguish the points where we have labels by calling them ``labeled real data'' and those without labels by calling them ``unlabeled real data''. Shown in figure~\ref{fig:real_images} are two images with their 9 spectral bands. 
The GAN generator supplies a fake image for every real image so that the WGAN-GP algorithm \cite{Gulrajani17} can be followed. Finally, once the model is trained, the features from the last layer are extracted for the images that have ground truth labels associated with them. A linear model is built on top of these features using cross-validated ridge regression. The model predictions are then compared to the ground truth values and a Pearson Correlation Coefficient is calculated. This is another metric by which we evaluate the performance of our models. In the work we have done so far, we have restricted our attention to the DHS survey taken in the year 2011 in Uganda.  This survey contains only 400 data points corresponding to ``geographic village cluster" level AWI labels. 

\section{Methods} \label{section:Methods}
\begin{figure}
    \centering
    \includegraphics[width=\linewidth]{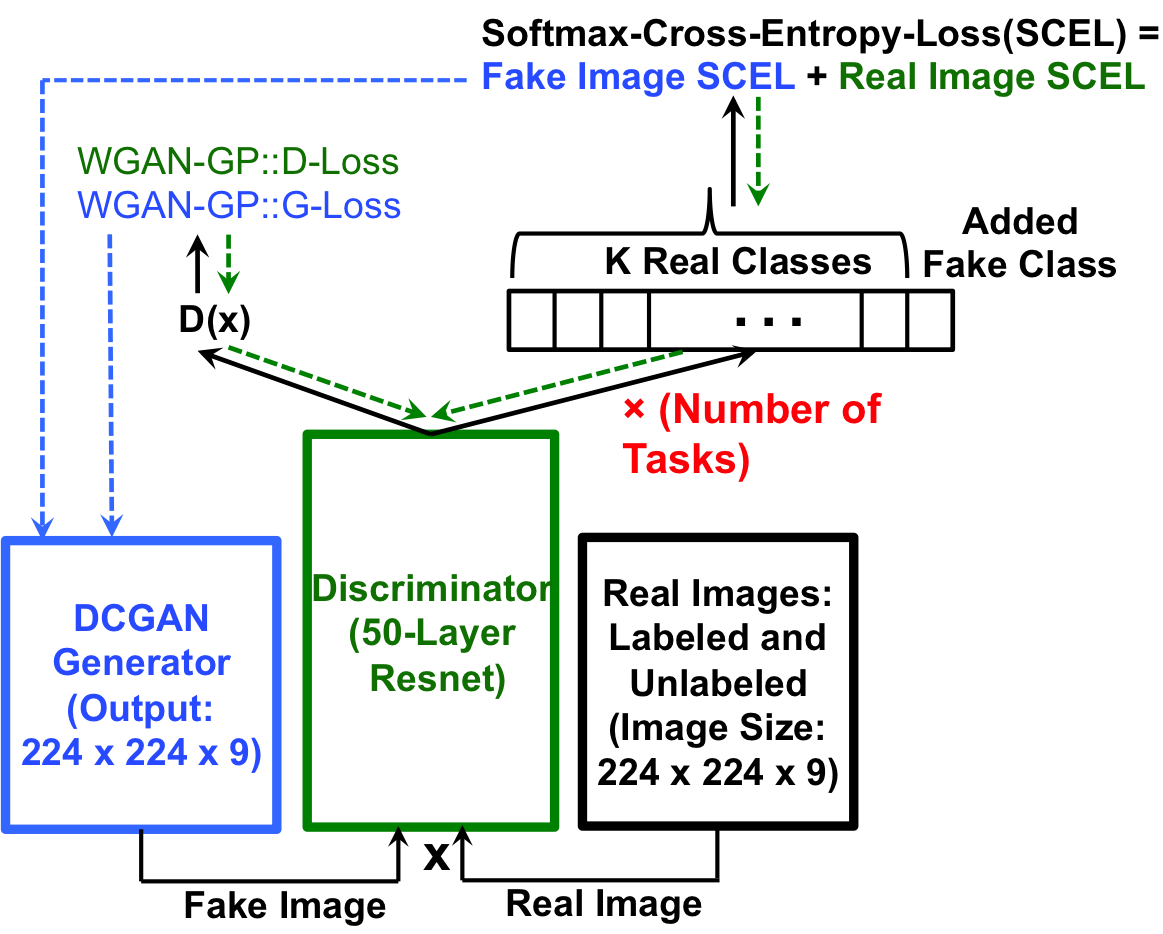}
    \caption{The WGAN Training Pipeline. The generator is a deep convolutional generator while the discriminator is a 50-layer ResNet. Solid lines show forward (inference) pass while the dashed lines show the backpropagation path. \label{fig:wgan}}
\end{figure}

As pointed out in \cite{Salimans16}, GANs \cite{Goodfellow14} boost the learning power of a semi-supervised model. This is implemented by having a generator supply ``fake images'' and the discriminator being modified such that if the classification task is associated with K classes, the model now has to predict (K+1) classes - K of them associated with real classes of interest while the (K + 1)th class represents the case where the provided image is fake.  We extend the loss proposed by \cite{Salimans16} to fit into a multitask paradigm. In the multitask framework, instead of having one classification task, we have multiple tasks, each of which is predicted by the model using a fully connected layer that stems from a shared model body.  Thus we append an extra class to the set of classes corresponding to each task, as shown in Figure~\ref{fig:wgan}.  However, vanilla GANs can be unstable during training \cite{Arjovsky17}.  It has been shown that Wasserstein GANs \cite{Arjovsky17} are a more stable variant of GANs.  Thus in addition to adding an extra class to each task, an extra task is added that corresponds to the WGAN task of predicting whether the input image is real or fake.  The multitask loss is the sum of each of these losses. Thus, the multitask loss function is minimized when the model (i) can predict the class association to be (K+1)-th class on all tasks and predict that the image is fake in the added WGAN task, if the provided image is fake, (ii) can associate a real labeled image with the correct label for the image in each task and predict that the image is real in the added WGAN task, and (iii) provided an unlabeled real image, it predicts that the image is not in the (K+1)-th class on all tasks and predicts that the image is real in the WGAN task. Here we use the word ``predict" to loosely mean estimation of some property that is not directly observed, rather than its common meaning of inferring something about the future.  Note that the discriminator in the WGAN task computes the Earth Mover Distance (Wasserstein-1 Distance or EMD) between the  distribution of real images and the distribution of images generated by the generator and thus is not ``predicting" a class in the conventional sense.  Recently, \cite{Gulrajani17} introduced the idea of enforcing a soft version of the 1-Lipschitz constraint required to minimize the WGAN value function constructed using the Kantorovich-Rubinstein duality.  This circumvents having to use weight clipping originally proposed by \cite{Arjovsky17} which was shown to be problematic \cite{Gulrajani17}.  For the WGAN task, we implement Algorithm 1 from \cite{Gulrajani17} by computing $D(x)$ from a fully connected layer that again stems from the shared model body. 

Thus for our semi-supervised tasks, the loss function incorporates labeled and unlabeled real data and WGAN generated fake data using the Kullbeck-Liebler divergence while the WGAN unsupervised task incorporates unlabled real data and WGAN generated fake data using the EMD.  To overcome the difference in the scales of our losses, we scale the WGAN loss with the hyperparameter $\alpha$ so that the scales of the two contributions to the total discriminator loss are similar. Summarizing the algorithm, the loss functions for the discriminator and the generator are:
\begin{align*}
& \;\;\;\;\;\; \;\;\;\;\;\; \;\;\;\;\;\; \text{Discriminator:}\\
\hat{x} \,\, \text{linear} &\,\, \epsilon \sim U[0,1] \,\, \text{superposition of real \& fake image}\\
L_{D_{\text{WGAN}}} &= \mathbb{E}_{z \sim P_z} \left[D(G(z))\right] - \mathbb{E}_{x \sim P_r} [D(x)] \\
&+ \lambda \mathbb{E}_{\hat{x} \sim P_{\hat{x}}} [(||\nabla_{\hat{x}} D(\hat{x})||_2 - 1)^2] \\
L_{D_t} &= - \mathbb{E}_{x, y \sim P_t} [\log \; p_{\text{model}_t}(y | x, y < K_t + 1)] \\
&- \mathbb{E}_{x \sim P_r} [\log( 1 - p_{\text{model}_t}(y = K_t + 1 | x) ) ] \\
&- \mathbb{E}_{z \sim P_z} [\log \; p_{\text{model}_t}(y = K_t + 1 | G(z)) ]\\
L_{D_{\text{multitask}}} &= \sum_{t \in \text{Tasks}} w_t L_{D_t}\\ 
L_D &= \alpha L_{D_{\text{WGAN}}} + L_{D_{\text{multitask}}} \\
& \;\;\;\;\;\; \;\;\;\;\;\; \;\;\;\;\;\; \text{Generator:}\\
L_{G_{\text{WGAN}}} &= - \mathbb{E}_{z \sim P_z} [D(G(z))] \\
L_{G_{\text{multitask}}} &= \sum_{t \in \text{Tasks}} w_t \; \mathbb{E}_{z \sim P_z} [log \; p_{\text{model}_t} (y = K_t + 1 | G(z) )] \\
L_G &= \alpha L_{G_{\text{WGAN}}} + L_{G_{\text{multitask}}}
\end{align*}

\begin{figure}
    \centering
    \includegraphics[width=\linewidth]{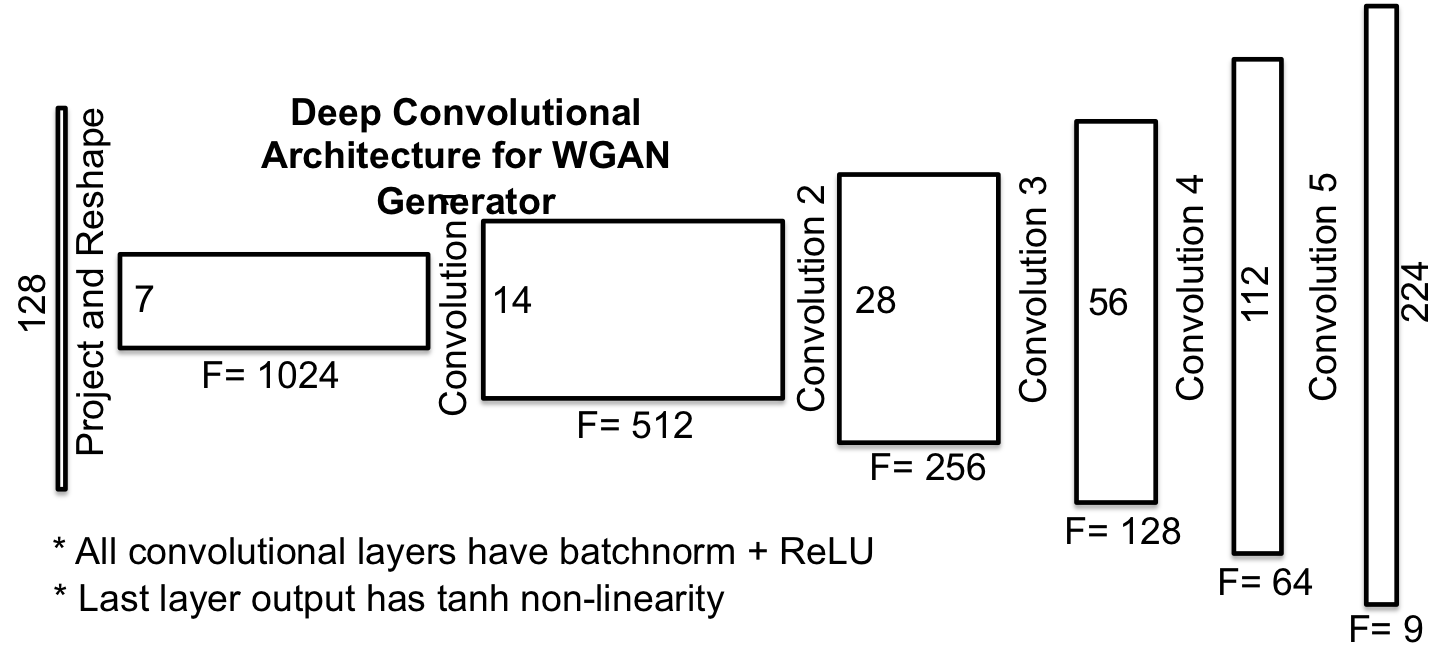}
    \caption{DCGAN \cite{DBLP:journals/corr/RadfordMC15} inspired deep convolutional generator. Each convolutional layer is followed by a batchnorm layer and a ReLU non-linearity. The last layer output has a tanh nonlinearity.\label{fig:gen}}
\end{figure}

The entire training pipeline along with which losses are used in updating and back-propagating into the model can be summarized in figure~\ref{fig:wgan}. The dashed lines show the loss functions that affect the model weight updates via back-propagation.
The generator used is inspired from the DCGAN deep convolutional generator \cite{DBLP:journals/corr/RadfordMC15}. The architecture we use for the generator is shown in figure ~\ref{fig:gen}. The output of this generator is set to be 224 x 224 x 9. The discriminator is the 50-layer ResNet from \cite{DBLP:journals/corr/HeZRS15}.  More specifically, the shared model body mentioned previously is composed of all layers of the 50-layer ResNet model from \cite{DBLP:journals/corr/HeZRS15} up to the average pooling layer.  The output of the average pooling layer is fed through a fully connected layer for each task as mentioned above.  Furthermore, when we refer to extracting features in the subsequent section, the features are exactly the output of the average pooling layer.

While training the discriminator, we monitor the training and validation accuracies of the semi-supervised tasks. Once the discriminator has been trained, we extract the last layer of the shared model body and use this as a feature representation of an input image. With these features, we train a linear ridge regression model on the locations where DHS survey labels for AWI are available. In this study, we focus on the Uganda for the year 2011. The regression is performed using a doubly nested cross-validation approach.  The inner loop of cross validation is used to search for the best weight on the regularization term (a hyperparameter) in ridge regression. That weight is used to make predictions for the test set in the outer loop of the doubly nested cross validation. Thus, at the end of each outer loop, a subset (specifically, the test set) of the datapoints are associated with a model prediction based on a regression over the other four folds. The predictions across all folds are shown in figure~\ref{a} for the experiments described in the next section. 

%There are two experiments we perform regarding the weights of the ResNet layers.

\section{Experiments and Results}
Our efforts so far have been focused on replicating the results by \cite{Jean16}.  To this end, we have created a pipeline that allows us to train a model based off the 50-layer ResNet described in \cite{DBLP:journals/corr/HeZRS15} to predict binned night light categories.  This is followed by a process that extracts features from the ResNet model for each location corresponding to a cluster in the Uganda DHS survey.  Finally we train a simple linear regression model from those features to predict the AWI score.  In order to replicate the transfer learning approach used by \cite{Jean16}, we have divided night lights into the same bins. The primary differences between our work and the work in \cite{Jean16} are the locations of the training data imagery, the model used, and the number of wavelengths of light captured in the imagery.
\begin{figure}
    \centering
    \subfloat[][]{
    	\includegraphics[width=0.45\linewidth]{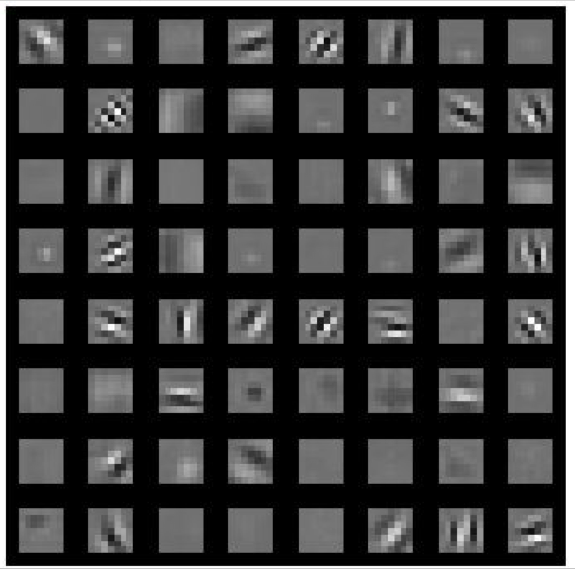}
    } \hfill
    \subfloat[][]{
    	\includegraphics[width=0.45\linewidth]{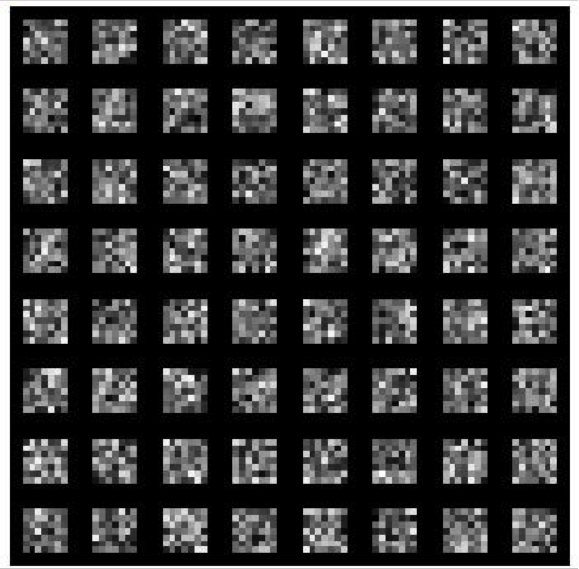}
    } \hfill
    \caption{On the left in figure (a) are shown the 64 first layer conv filters for the NIR channel when the initialization of the weights is ``same init'' meaning that the weights in each channel is the mean of the RGB channels of the pretrained ResNet weights. On the right in figure (b) are shown the same filters at convergence when the initialization of the weights is ``random init'' meaning that the weights are chosen from a truncated normal distribution with mean and standard deviation of the pretrained ResNet weights. \label{resnet_weights}}
\end{figure}
\begin{figure}
    \centering
    \includegraphics[width=0.5\textwidth]{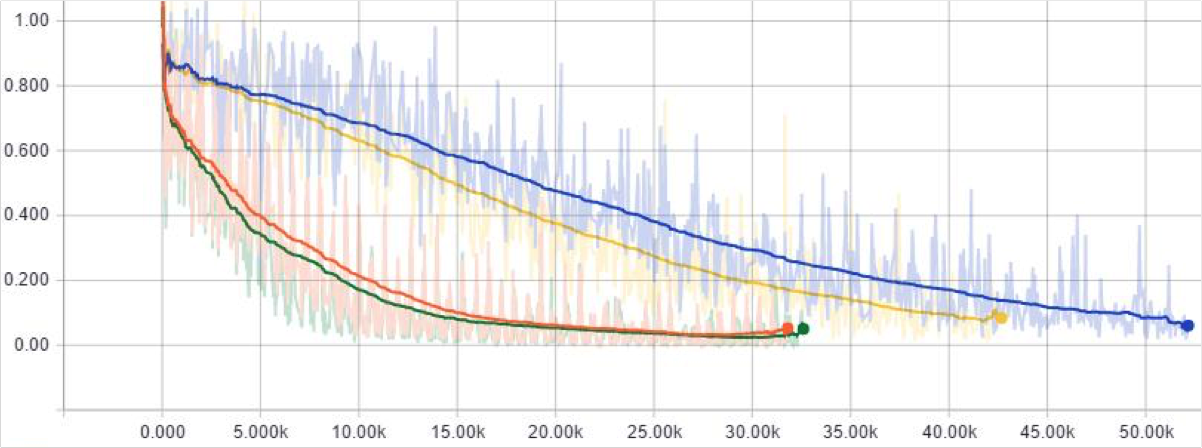}
    \caption{The plot shows the convergence behavior of the loss function of the discriminator for four experiments: (Orange) Use ``Around DHS dataset" with ``same init'' initialization scheme; (Green) Use only the RGB channels of the ``Around DHS dataset'' with ``same init'' initialization scheme; (Blue) Use ``All Africa dataset'' with ``random init'' initialiation scheme; and (Yellow) Use ``All Africa dataset'' with ``same init'' initialization scheme.   \label{e}}
\end{figure}

% Experiment description
We have conducted several experiments regarding the high level hyperparameters of our model which include the initialization of our model, the dataset our model is trained with, and the number of Landsat 7 channels supplied to the model.  Note that because our imagery is multispectral (and thus has 9 channels), the size of the filters in the first layer of our ResNet model have changed relative to their Imagenet counterparts.  As such, initialization directly from Imagenet parameters is not possible for the network.  We define two initialization schemes, both of which initialize all weights in the second layer through the last layer to the corresponding Imagenet weights.  The first initialization scheme, which we denote ``random init'', initializes the RGB components of the filters in the first layer to their Imagenet values, and then initializes the other channel's components randomly using a truncated normal distribution with the mean and standard deviation of the weights in the first layer of the ResNet model pretrained on Imagenet .  The second initialization scheme, which we denote ``same init'', again initializes the RGB components of the filters in the first layer to their Imagenet values, but now initializes the other channel's components with the mean of the pretrained Imagenet weights taken across the RGB axis.  In other words, for every position in each first layer filter, its six non-RGB channel weights are initialized as the mean of the three RGB channel weights available from the pretrained Imagenet model.  Additionally, we define two datasets over which we conduct experiments.  These datasets are described in more detail in Section~\ref{section:Data}.  The first dataset is denoted the ``All Africa dataset'' and was sampled uniformly from an evenly spaced grid over the African continent.  This dataset is visualized in Figure~\ref{old_locs}.  The second dataset is denoted the ``Around DHS dataset'' and was sampled such that its location distribution is similar to the location distribution from which the DHS surveys are sampled.  This dataset is visualized in Figure~\ref{new_locs}. Finally we experiment with whether including only the Blue, Green, and Red channels rather than all nine Landsat 7 channels degrades or improves the performance of the model.

%  low level hyper parameters
A manual hyperparameter search was done over the learning rate, batch size, and learning rate decay. Figure~\ref{e} shows the training loss of each of our experiments using their final hyperparameter settings.  All models use a batch size of 115, which is the maximum our hardware would allow.  Additionally, all models feature $l_2$ weight decay on all layers with a regularization strength of 0.00004.  Our best model uses a learning rate of 0.01 and a learning rate decay of 0.98 per epoch.  We additionally lower the learning rate by a factor of 5 after 25 epochs of training as this is where the model plateaus.  The best performing model is initialized using the ``same init'' initialization scheme, uses only the RGB channels from the Landsat 7 imagery, and is trained on the ``Around DHS dataset''.

% Evaluation 
We evaluate our methods on the AWI available in DHS survey sampled in Uganda in the year 2011.  First, predictions are made for all locations with data available using the ridge regression model trained as described in Section~\ref{section:Methods}.  We then evaluate the success of our model by computing the Pearson correlation coefficient between the predicted AWI values and the actual AWI values.  Reporting the Pearson correlation coefficient allows for a fair comparison between this work and the work in \cite{Jean16}.  As a tool to understand our methods, the night light training and validation prediction accuracies are reported as well.
\begin{figure}
    \centering
    \includegraphics[width=\linewidth]{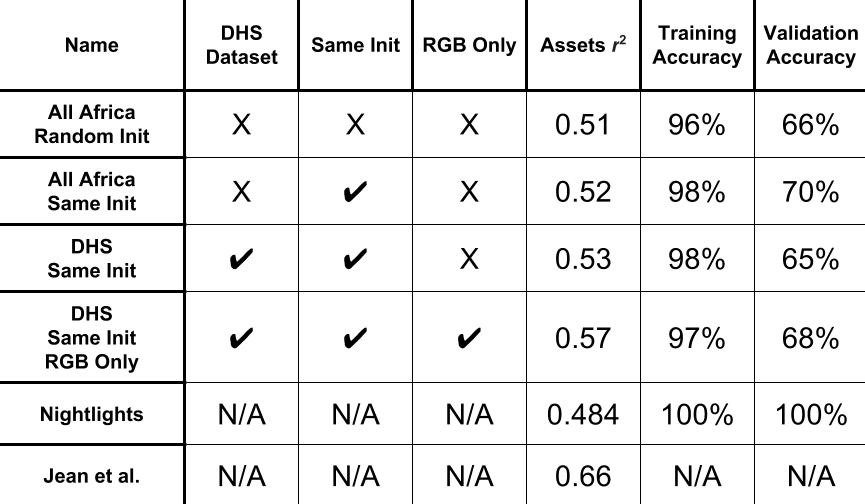}
    \caption{A table summarizing the results of our experiments. Names in the first four rows refer to experiment qualities. Nightlights refers to using the raw nightlight values.  Jean et al. refers to \cite{Jean16}.  The first two columns highlight the initialization scheme and the dataset used. The third column denotes if all the 9 channels were used or only the first 3 channels were used. The fourth column denotes the $r^2$ obtained in the linear model when the features extracted by the discriminator are used. The last two columns are the training accuracies and validation accuracies for the discriminator models.  \label{t}}
\end{figure}

\begin{figure}
    \centering
    \includegraphics[width=0.38\textwidth]{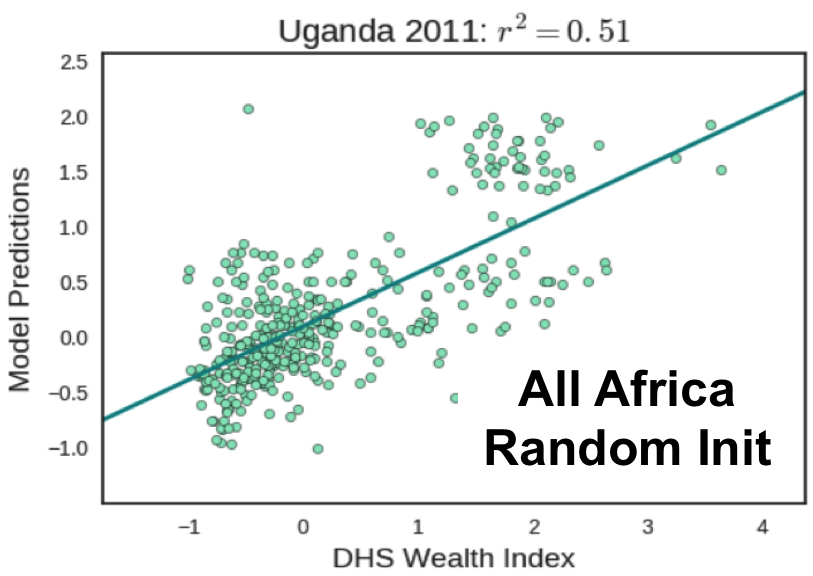}
    \includegraphics[width=0.38\textwidth]{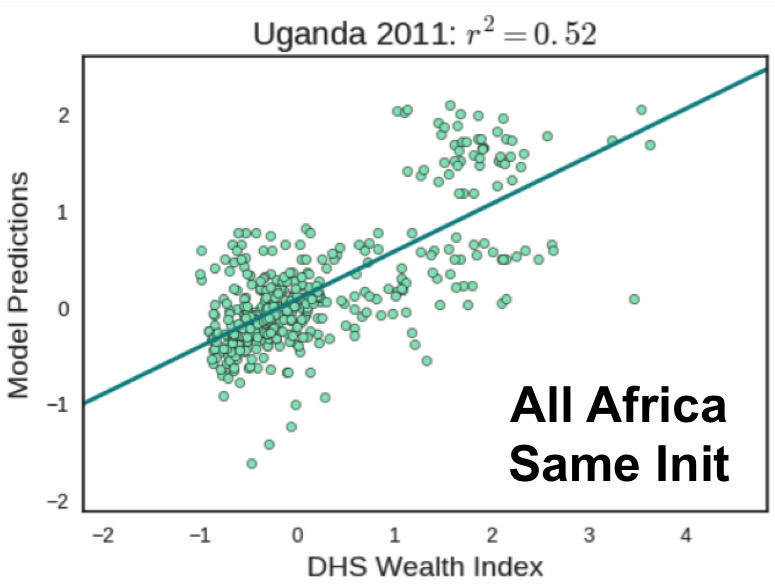}
    \includegraphics[width=0.38\textwidth]{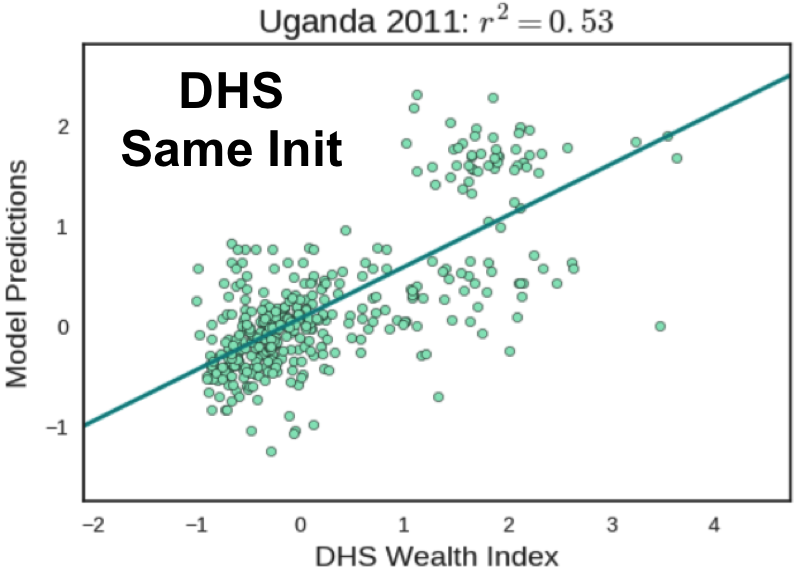}
    \includegraphics[width=0.38\textwidth]{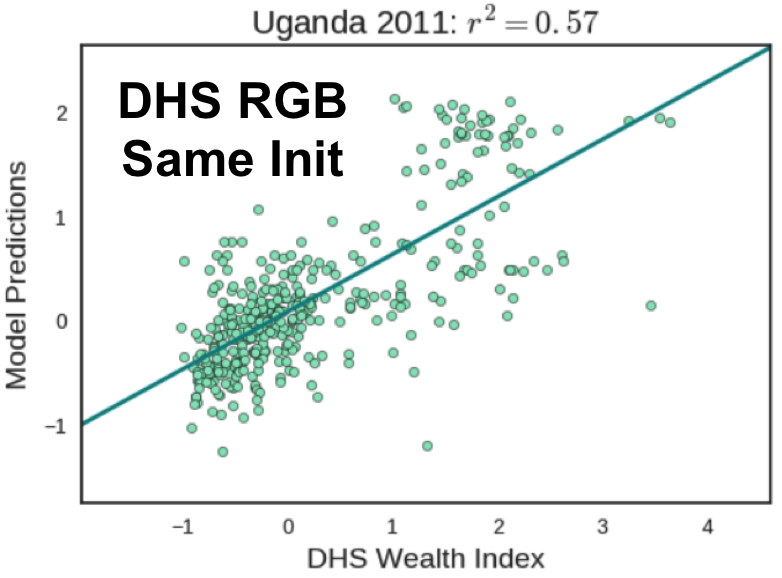}
    \caption{Figures show the relation between the model predictions and the ground truth values of AWI in Uganda for the year 2011. The line demonstrates the linear ridge regression model that is built on top of the predictions.  Asset values are predicted using the doubly nested cross validation strategy described in Section~\ref{section:Methods}. \label{a}}
\end{figure}

% Results
Figure~\ref{t} reports the results of the experiments conducted while Figure~\ref{a} compares each model's predicted AWI to the ground truth data in the 2011 DHS survey in Uganda.  By comparing each model's night light prediction accuracy on the training and validation sets we can see signs of overfitting on the training data, as the training accuracy is much higher than the validation accuracy.  While regularization would be beneficial, incorporating additional tasks may also be sufficient to reduce overfitting and would likely result in improved generalization of the features extracted by the model \cite{Caruana97}.  Sadly we have not surpassed the results of our ``oracle'' \cite{Jean16}, who achieved a correlation coefficient of approximately 0.66 (we have achieved 0.57).  However, even with overfitting we do surpass the baseline correlation coefficient computed between night light values and the AWI, which is 0.484.
% Maybe include a plot of the validation accuracy if it helps make the overfitting point?

% Results cont. commenting on the three experimental axis
As one may expect, using the ``Around DHS dataset'' and the ``same init'' initialization resulted in increase model performance.  Using the ``Around DHS dataset'' shifts the model's training distribution closer to the distribution of images in which DHS AWI assets are available which reduces the problem of covariate shift \cite{Shimodaira2000227}.  Likewise, using the ``same init'' allows the model to retain the information present in the pretrained weights and may better pair the non-RGB filter weights with the pretrained RGB weights being used in both initialization schemes.  In order to better understand the effects of the different initialization schemes, the weights of the first layer filters corresponding to the Near Infrared (NIR) channel are visualized in Figure~\ref{resnet_weights}.  It seems that when the ``same init'' is used, the model retains filters that look similar to those of the pretrained ImageNet model.  Unexpectedly, removing the non-visible channels from the input imagery results in a better performing model.  This may be explained by the model's propensity to overfit, as the model may be fitting to noise in the additional image channels.  We expect that when overfitting is resolved, either by adding additional tasks or incorporating stronger regularization, that using all Landsat 7 image bands would result in a more powerful model.

\section{Conclusions and Future Work}
We have proposed a technique to extend the semi-supervised learning method proposed in \cite{Salimans16} to a multitask setting.  This is done by adding an extra class corresponding to fake images to each semi-supervised task.  We also incorporate the WGAN loss proposed in \cite{Gulrajani17} in order to achieve greater stability in training our generator.  This involves adding an additional task whose loss is described in Algorithm 1 in \cite{Gulrajani17}.  We operate in a domain where only 5\% of the data is labeled while 95\% of the data is unlabeled.  In domains where labels are scarce, the multitask WGAN model proposed can be used for simultaneously predicting the output of several semi-supervised tasks.  As of now, the discriminator is capable of over-fitting to the data which demonstrates that the model has sufficient capacity. As more tasks are added, they should have a regularizing effect on the model and improve the generalization of our model's features which is one of our next steps. Our model's predictions comfortably achieve a higher correlation coefficient with ground truth AWI than raw night lights do.  We consider outperforming night lights to be a reasonable ``baseline'' for the model's success.  We are still a small distance away from overcoming the 0.66 correlation coefficient achieved by \cite{Jean16} which is our ``oracle''. The WGAN generator has proven very hard to train. Training the generator to convergence and getting fake images that mimic actual satellite images is our next goal.

\clearpage
\section{Acknowledgements}
We would like to thank Stefano Ermon, Volodymyr Kuleshov\footnote{Ph.D. Candidate, Department of Compute Science}, Marshall Burke, and David Lobell for providing algorithmic and domain insights as well as constructive criticism.  We would like to thank George Azzari and Matthew Davis\footnote{Center on Food Security and the Environment, Stanford University, Stanford, CA, USA.} for helping us gather and understand the satellite and economic data.  We would like to thank Neal Jean\footnote{Ph.D. Candidate, Department of Electrical Engineering} for both fruitful discussion as well as providing the code used to train the ridge regression model from features extracted by our Resnet-based model to predict poverty.  This assisted in allowing us to compare our results accurately.  We would like to thank \cite{resnet_implementation} for providing starting code for our Resnet architecture as well as \cite{Gulrajani17} for providing the code from which our DCGAN generator was derived.  Likewise, we would like to thank \cite{Salimans16} for making public the code on which our implementation of their semi-supervised loss is based.  Our work is implemented using Tensorflow \cite{tensorflow2015-whitepaper} and our data was gathered using Google Earth Engine \cite{google_earth_Engine}.

{\small
\bibliographystyle{ieee}
\bibliography{report}
}

\end{document}